\documentclass[conference]{IEEEtran}
\IEEEoverridecommandlockouts
\usepackage[l2tabu,orthodox]{nag}

\usepackage[
    backend=bibtex8,
    style=ieee,
    sorting=none,
    natbib=true,
    doi=false,
    isbn=false,
    url=false,
    eprint=false,
    maxcitenames=1,
    mincitenames=1
]{biblatex}

\usepackage[pdftex,colorlinks]{hyperref}
\usepackage[printonlyused]{acronym}

\usepackage{siunitx}
\usepackage[all]{nowidow}

\usepackage[dvipsnames]{xcolor}

\usepackage{lipsum}

\usepackage{xspace} %

\usepackage[pdftex]{graphicx}

\usepackage{epstopdf}

\usepackage{import}

\graphicspath{{./latexGoodPractices/}}

\usepackage{booktabs}

\usepackage{tabularx}
\usepackage{multirow, multicol}

\usepackage{amssymb,amsfonts,amsmath,amscd}

\usepackage{bm}

\newcommand{\bbm}{\begin{bmatrix}}
\newcommand{\ebm}{\end{bmatrix}}

\usepackage{fancyhdr}
\fancypagestyle{withfooter}{
  
  \fancyhead[L]{}
  \fancyhead[R]{}
  \fancyfoot[C]{\footnotesize Accepted to the IEEE ICRA Workshop on Field Robotics 2024}
}

\acrodef{SLAM}{simultaneous localization and mapping}
\acrodef{SOTA}{state-of-the-art}
\acrodef{SSMR}{skid-steering mobile robot}
\acrodef{AMR}{Ackermann mobile robot}
\acrodef{UGV}{uncrewed ground vehicle} 
\acrodef{IDD}{ideal differential-drive}
\acrodef{ICR}{instantaneous center or rotation}
\acrodef{RTK}{Realtime Kinematics}
\acrodef{GNSS}{Global Navigation Satellite System}
\acrodef{ROC}{radius of curvature}
\acrodef{IMU}{inertial measurement unit}
\acrodef{MPC}{model predictive control}
\acrodef{GP}{Gaussian processe}
\acrodef{BLR}{Bayesian linear regression}
\acrodef{IPEM}{integrated prediction error minimization}
\acrodef{MLP}{multilayer perceptron}
\acrodef{ICP}{iterative closest point}
\acrodef{MRMSE}{multi-step root mean squared error}
\acrodef{T-MRMSE}{translational multi-step root mean squared error}
\acrodef{R-MRMSE}{rotational multi-step root mean squared error}
\acrodef{M-Z-score}{multi-step Z-score}
\acrodef{DRIVE}{Data-driven Robot Input Vector Exploration}
\acrodef{VISTA}{Vehicle Input Space Training Assistant}
\acrodef{ATV}{all-terrain vehicle}
\acrodef{ICR}{instantaneous centre of rotation}

\newcommand{\CMDBODYVEL}{^{\robotf}\bm{f}}
\newcommand{\SLIPBODYVEL}{^{\robotf}\bm{g}}

\newcommand{\INPUTVECTOR}{\bm{u}}

\newcommand{\robotf}{\mathcal{R}} %

\begin{document}

\title{Comparing Motion Distortion Between Vehicle Field Deployments}

\author{Nicolas Samson$^{1}$, Dominic Baril, Julien Lépine, François Pomerleau$^{1}$
\thanks{*This research was supported by the Fonds de recherche du Québec – Nature et technologies (FRQNT) and by the Natural Sciences and Engineering Research Council of Canada (NSERC) through grant CRDPJ 527642-18 SNOW (Self-driving Navigation Optimized for Winter).}%
	\thanks{$^{1}$Northern Robotics Laboratory, Université Laval, Quebec City, Quebec, Canada
		{\texttt{\small \{nicolas.samson francois.pomerleau \}@norlab.ulaval.ca}}}%
}

\maketitle
\thispagestyle{withfooter}
\begin{abstract}
Recent advances in autonomous driving for \acp{UGV} have spurred significant development, particularly in challenging terrains. 
This paper introduces a classification system assessing various \ac{UGV} deployments reported in the literature. 
Our approach considers motion distortion features that include internal \ac{UGV} features, such as mass and speed, and external features, such as terrain complexity, which all influence the efficiency of models and navigation systems.
We present results that map \ac{UGV} deployments relative to vehicle kinetic energy and terrain complexity, providing insights into the level of complexity and risk associated with different operational environments. 
Additionally, we propose a motion distortion metric to assess \ac{UGV} navigation performance that does not require an explicit quantification of motion distortion features. 
Using this metric, we conduct a case study to illustrate the impact of motion distortion features on modeling accuracy.
This research advocates for creating a comprehensive database containing many different motion distortion features, which would contribute to advancing the understanding of autonomous driving capabilities in rough conditions and provide a validation framework for future developments in \ac{UGV} navigation systems.
\end{abstract}


\acresetall

\section{Introduction}
Since motion models are a fundamental component of autonomous navigation systems, improving their accuracy and robustness to complex motion is critical to enabling efficient autonomy in various scenarios. 
Much attention has been given to model formulation in the last years~\cite{Teji2023}, however, motion complexity in evaluation datasets varies greatly.   
While we recently proposed a protocol to standardize model training and evaluation dataset gathering~\citep{Baril2024}, no standard metric exists for comparing model performances on different \ac{UGV} terrains.  
Such a metric is key to enabling efficient comparisons between models and easing the selection of motion models based on operational requirements. 
\begin{figure}[htbp]
    \centering
    \includegraphics[width=0.5\textwidth]{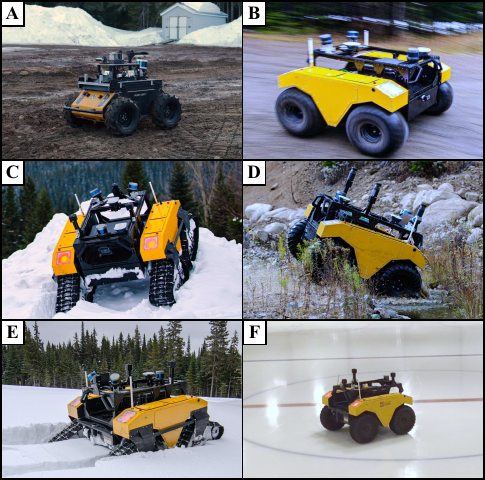}
    \caption{Examples of motion distortion features that affect \ac{UGV} motion.
            Heavier vehicles with
            higher top speeds have more inertia (A, B). 
            Aggressive motion leads to transitory behavior and non-linear dynamics as opposed to slower, conservative driving (A, B). Terrain steepness and roughness also lead to slippage and loss of contact between wheels and terrain (C, D). 
            Lastly, terrain hardness and friction significantly impact vehicle motion (E, F). 
            In F, the Warthog is navigating on a resurfaced ice rink.}
    \label{fig:intro}
\end{figure}
A vast amount of factors affect~\ac{UGV} motion to a varying degree. A few examples can be the \ac{UGV} weight and speed, the terrain steepness, roughness, hardness, and friction coefficient, as shown in~\autoref{fig:intro}. 
In this work, we use the terms \emph{motion distortion features} for these factors.
We regrouped them into two categories: internal and external. 
Internal motion distortion features are related to the driving speed, aggressive maneuvers, or vehicle properties.
External motion distortion features are related to terrain properties, such as steepness, hardness, or the friction coefficient between the terrain and the tire. 

Quantifying terrain properties and their impact on the \ac{UGV} motion is complex and challenging~\cite{Seegmiller2016}. 
Methods based on exteroceptive sensors have the disadvantage of misinterpretation when the ground is composed of layers of different materials because hidden layers change the terrain properties but are not necessarily perceptible by the sensors.
For example, a thin layer of snow can hide from a camera an icy terrain that is more slippery than snow. 
Methods based on proprioceptive sensors have the inconvenience that the mobile robot needs to cross the terrain to obtain its properties, which can put the robots in precarious terrain that causes immobilization~\cite{Teji2023}. 

In this work, we propose a metric that solves the problem of comparing motion model performances that have not been tested in the presence of the same motion distortion features.
Our approach consists of subtracting observed velocity from the ideal slip-less velocity computed by an ideal nominal model.  
The underlying hypothesis is that the ideal motion model assumes a pure rolling motion, which occurs only at low speed on simple and flat terrain. 
The ideal motion model error is a measurement of the dynamics imposed by the combination of the terrain, the trajectory, and the vehicle. 
The minimal computational requirement of our approach facilitates its use for comparing model evaluation conditions, at the cost of not being able to discern the difficulty caused by the internal motion distortion from the external ones.

In the first section of this article, we list and describe the motion distortion features in recent motion modeling and control papers.
These papers are also mapped based on the concept of external and internal motion distortion features to show the need for a metric agnostic to the source of motion distortion. 
In the second section, we present the mathematical formulation for our proposed motion difficulty metric. 
In the last section, we present a case study of the metric on four datasets containing two types of vehicles deployed respectively on two terrains. 

\section{Related work on \ac{UGV} motion distortion features}

\begin{figure}[h!]
    \centering
    \includegraphics[width=0.5\textwidth]{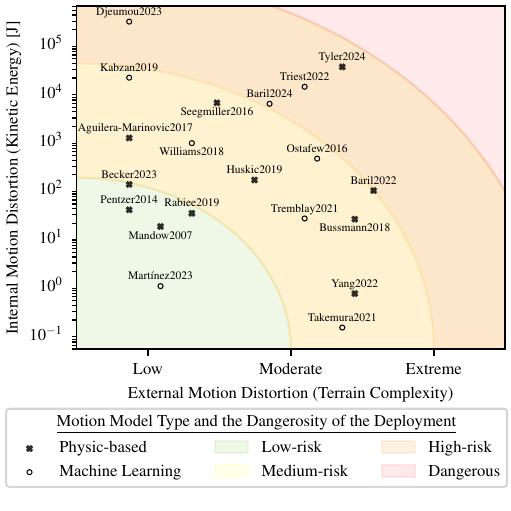}
    \caption{Qualitative mapping of the related work based on the motion distortion features respectively represented by the maximum kinetic energy of the vehicles used in the experiment and the most complex terrain used in the experiment. 
    The type of motion model developed by the authors is also presented using different markers. 
    The area created by the internal and external motion distortion features is also divided based on the risk associated with mobile robot deployment in these conditions.
    One should note the logarithm scale used for the \textit{y}-axis.}
    \label{fig:RW_qualitative}
\end{figure}

We divide motion distortion features into two categories: internal and external.
Internal motion distortion features include factors that are caused by the vehicle properties or a command vector~$\INPUTVECTOR$. 
For example, commanding the \ac{UGV} to drift will generate more internal motion distortion than turning slowly without skidding.  
External motion distortion features include factors caused by the environment, such as terrain properties. 
In this section, we provide a comprehensive list of both types of motion distortion features, as observed in previous~\ac{UGV} field deployments from different authors. 
In~\autoref{fig:RW_qualitative}, these deployments are also qualitatively mapped based on the description of the article's experiment, with external and internal motion distortion features represented on different axes.
The information on the motion distortion features varies from paper to paper. 
Consequently, the internal motion distortion features are approximated by the maximum kinetic energy of the heaviest \ac{UGV} used in the experiment presented in each article. 
This approximation is used because it gives the order of energy engaged in the system.
Therefore, the maximum kinetic energy indicates an approximation of the order of the forces that must be applied to change the state of the \ac{UGV}.
Another benefit of approximating the internal motion distortion in the mapping by the maximum kinetic energy is that it can be computed for more articles because it only depends on two commonly given parameters: the mass and the square of the vehicle maximum speed. 
In \autoref{fig:RW_qualitative}, the external motion distortion features are approximated by the most complex terrain tested in each article.
The terrain complexity has been qualitatively ordered based on the information available about the terrain steepness, roughness, hardness, and friction coefficient. The terrain complexity goes from flat asphalt to soft soil like sand and deep snow with steep slopes~\cite{Yang2022, Baril2022}.  
In the worst case, terrain information was inferred based on the visual information present in the article.

\textbf{Internal motion distortion features} are related to the vehicle and how it is commanded.   
\citet{Rabiee2019} quantify the accuracy improvement when including the impact of vehicle inertia in the motion model. 
\citet{Kabzan2019} show that high-speed driving on a racetrack with an autonomous race car leads to tire saturation, causing motion distortion, which they model with \acp{GP}.
\citet{Djeumou2023} combine this high-speed driving with aggressive motions, pushing the tires past the saturation point.
In their work, they successfully perform autonomous drifting with a full-size car by modeling tire forces with a neural network. The car weight (\SI{1450}{kg}) and top speed (\SI{13.86}{m/s}) make it the motion model validated with the highest kinetic energy.
\citet{Becker2023} also used tire force models to improve autonomous racing performance for a 1:10 scale racecar.
Aggressive driving was also performed on a dirt racetrack by \citet{Williams2018}, who modeled motion distortions with a neural network. However, both authors used a scaled-down car version, which results in lower kinetic energy.
In their work, they show that aggressive motion on dirt roads leads to a change in the vehicle orientation, which affects the resulting body velocity. 
The steering geometry also has an impact on the internal motion distortion features.
For example, Ackermann steering geometries are designed to minimize wheel skidding, while~\acp{SSMR} require skidding for angular motion~\citep{Mandow2007}. 
\citet{Pentzer2014} showed that the inherent skidding and slipping of \ac{SSMR}s increased with the speed of the \ac{SSMR}.

\textbf{External motion distortion features} are related to the surrounding environment and the vehicle interaction with that environment. 
A small change of terrain properties between the ground at calibration and the ground at deployments can lead to a divergent motion model if the parameters are not updated continuously~\cite{Bussmann2018}. %
\textbf{Terrain steepness} has multiple impacts on the vehicle motion. 
\citet{Seegmiller2016} showed that the vehicle pitch and roll orientation affect motion to a scale depending on the terrain properties.
\citet{Ostafew2016} observed a similar phenomenon when driving a \ac{SSMR} over gravel piles.
\citet{Takemura2021} deployed a lunar rover on loose terrain with an inclination leading to ground deformation and terrain slip.
As part of their large off-road driving dataset, \citet{Triest2022} deployed an \ac{ATV} weighing approximately \SI{730}{kg} in sloped terrains and benchmarked learning approaches to account for the resulting motion distortion. 
To quantify the difficulty of their dataset, they measure the average variation of height per second.
Steep terrain also limits the ability of \ac{SSMR} to turn around itself because its inherent skidding combined with the slopes results in the \ac{SSMR} skidding towards the bottom of the slopes instead of skidding around itself~\citep{Martinez2023}.  %

\textbf{Terrain roughness} can affect vehicle motion by reducing the contact between the ground and the wheels. A rougher terrain leads to a higher number of contact losses and, therefore, a more complex motion. 
\citet{Yang2022} explain that a wrong estimation of the contact area between the deformable soil and the wheels of a \ac{UGV} can lead to significant motion prediction error and the failure of the dynamic-motion-model simulation. 
Similarly, \citet{Tremblay2021} deployed a small-size~\ac{SSMR} in boreal forest trails cluttered with fallen branches and roots that modify the contact between the ground and the wheels.
In their work, they learn motion distortions based on multiple sensor modalities. 
In parallel, \citet{Aguilera-Marinovic2017} developed another general physic-based motion model approach that can estimate the impact of the loss of contact between the ground and the wheels on the movement of a mobile manipulator installed on a skid-steer. 

\textbf{Ground hardness and friction} are also important external motion distortion features. 
\citet{Fiset2021} studied the energy-loss and ground bulldozing effect for a~\ac{SSMR} turning in loose sand. 
In previous work, we show that deep snow navigation can lead to complete vehicle immobilization, which would represent an extreme case of motion distortion~\citep{Baril2022}.
As for terrain friction, \citet{Huskic2019} showed that the same set of parameters was sufficient to conduct high-speed path following on relatively uniform terrain, namely concrete, light grass, and gravel.
However, in previous work, we show that navigation on surfaced ice generates much higher motion distortion than on other hard ground~\citep{Baril2024}. %
In this case, we observed that the low friction force leads to longer transitory behavior and a high impact of vehicle inertia.
Even advanced slip learning-based models failed to predict motion accurately in this terrain \citep{Baril2024}. 

As shown in \autoref{fig:RW_qualitative}, motion models are often tested in the presence of different motion distortion features. This variety of motion distortion features makes it hard to compare the performances of motion models. 
For example, \citet{Yang2022} have a motion model that performed well in adverse external motion distortion features like non-cohesive sand and slopes but was only tested for low internal motion distortion features. 
\citet{Djeumou2023} did the opposite by validating its motion model only in the presence of high internal and low external motion distortion features. 
A motion distortion metric agnostic to the causes could help reduce the different motion distortion features in one dimension. 

The mapping in \autoref{fig:RW_qualitative} also lights a remaining research challenge: \textbf{the development of a motion model for high-risk deployments}. 
To the best knowledge of the authors, only \citet{Han2024,Triest2022,Djeumou2023} have pushed the motion validation in the high-risk validation by deploying a \ac{UGV} in complex terrain. 
\citet{Triest2022} presented a learning-based motion model that can predict the position of a Yamaha Viking navigating on a forest track with a median height variation of \SI{0.2253}{m/s} and a vehicle speed of up to \SI{6}{m/s}. 
Similarly, \citet{Han2024} deployed an autonomous \ac{ATV} that successfully navigated at a speed of up to \SI{10}{m/s} on different complex terrains like v-ditches, dried rivers, slopes of 20 degrees, and craters. 
However, both of these motion models have been validated with a respective maximum kinetic energy that represents only \SI{1.5}{\%} and \SI{0.11}{\%} of the kinetic energy used to validate the motion model of \citet{Djeumou2023}.   
This statistic proves that motion model for high-risk deployments on more complex terrain have not reached their maximum performances. 
In addition, some terrains remain out of reach. 
For example, \citet{Baril2022} deployed a \ac{SSMR} in snow and observed that deep snow is a highly complex terrain to navigate because the \ac{SSMR} tends to sink in it. 
The sinkage of the vehicle reduces the viable angular motion. If the turning restrictions are not respected, the \ac{SSMR} can sink to the point where it is impossible to recover without human intervention.

\section{Motion distortion metric}
\label{sec:metho_rw}
While we provide a comprehensive list of motion distortion features that are known to impact~\ac{UGV} motions, many others exist.
Quantifying all motion distortion features would be
costly because it would require the complete characterization of the UGV, the terrain, and the interaction between the two. 
The expensive material and expertise required to accomplish these tasks motivate the usage of our motion distortion metric, which only requires vehicle command and velocity. 

\begin{figure}[htpb]
    \centering
    \includegraphics[width=0.3\textwidth]{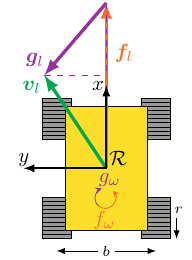}
    \caption{A 2D planar representation of our model formulation's important variables.
            The ideal motion~$\CMDBODYVEL$ is represented in orange.
            The estimated resulting body velocity~$^{\robotf}\bm{v}$ is represented in green.
            Motion distortion~$^{\robotf}\bm{g}$, or body-level slip, is shown in purple.
            Angular components for all velocities are represented as~$(\cdot)_\omega$. Linear components for all velocities are represented as~$(\cdot)_l$ 
            For the case of~\acp{SSMR}, ideal motion is parameterized by wheel radius~$r$ and vehicle width~$b$.}
    \label{fig:motion_model}
\end{figure}

To quantify the difficulty of a~\ac{UGV} motion dataset, we set basic concepts for the model formulation, which are illustrated in~\autoref{fig:motion_model}.
All the velocities are defined in the vehicle body frame $\robotf$, which has its origin located on the vehicle center of rotation and $x$-axis parallel to its longitudinal direction.

We proceed by estimating motion distortion~$\!\SLIPBODYVEL_{t}$ from an ideal driving model, assuming slip-less motion, similar as done in the previous work of \citet{Baril2024}:
\begin{equation}
	\!\SLIPBODYVEL_{t} = \!\CMDBODYVEL_t(\INPUTVECTOR_{t}) -  ^{\robotf}\bm{v}_{t},
	\label{eq:body_vel}
\end{equation} 
where~$t$ is the time at which the metric is calculated,~$\SLIPBODYVEL_{t}$ is the slip body velocity and~$^{\robotf}\bm{v}_{t}$ is the observed body velocity, estimated by a vehicle localization system.
Ideal motion is dependent on the vehicle command, or input vector~$\INPUTVECTOR_{t}$.
In this work, we provide model formulation for~\acp{SSMR}, for which ideal motion is computed by the ideal differential-drive model:
\begin{equation}
	\CMDBODYVEL_t(\INPUTVECTOR_{t}) = \begin{bmatrix}
		f_x \\
		f_y \\
		f_\omega \\
	\end{bmatrix} = 
	r \begin{bmatrix}
		\frac{1}{2}, \frac{1}{2} \\
		0, 0 \\
		-\frac{1}{b}, \frac{1}{b} \\
	\end{bmatrix} \begin{bmatrix}
		\omega_{l_t} \\
		\omega_{r_t} \\
	\end{bmatrix} ,
	\label{eq:cmd}
\end{equation}
where~$\omega_{l_t}$ and~$\omega_{r_t}$ are the left and right wheel velocities.
The model parameters are the vehicle wheel radius~$r$ and vehicle width~$b$, as shown in~\autoref{fig:motion_model}.
One could easily adapt this model formulation to Ackermann vehicles, by switching the differential-drive model with a bicycle model, as described by \citet{Becker2023}.

We propose a simple motion distortion metric consisting of the modulus of the slip vector ~$\left|\left|\SLIPBODYVEL_{t}\right|\right|$. 
This metric quantifies the difficulty of a motion without explicitly measuring the sources of the motion distortion. 
The metric is also straightforward to compute, only requiring a localization system to estimate the body frame displacement. 
Once all the moduli of the slip vectors are calculated, it is possible to compare the difficulties of datasets by doing hypothesis testing and verifying whether the difficulty levels of the two datasets are significantly different.

\section{Case study}
\label{sec:results}

The proposed metric was tested on four datasets collected with the \ac{DRIVE} protocol~\citep{Baril2024} to show that it can be used to evaluate the difficulty of the terrain, trajectory, and vehicle combination. 
Two datasets were collected with a \emph{Clearpath Husky A200} robot navigating separately on tile and on snow.
The \emph{Husky} has a maximum speed of \SI{1}{m/s} and weighs \SI{75}{kg}.
The two other datasets were separately collected with a \emph{Warthog} on ice and gravel. 
The \emph{Warthog} weighs \SI{470}{kg} and has a maximum speed of \SI{5}{m/s}. 

\begin{figure}[h!]
    \centering
    \includegraphics[width=0.49\textwidth]{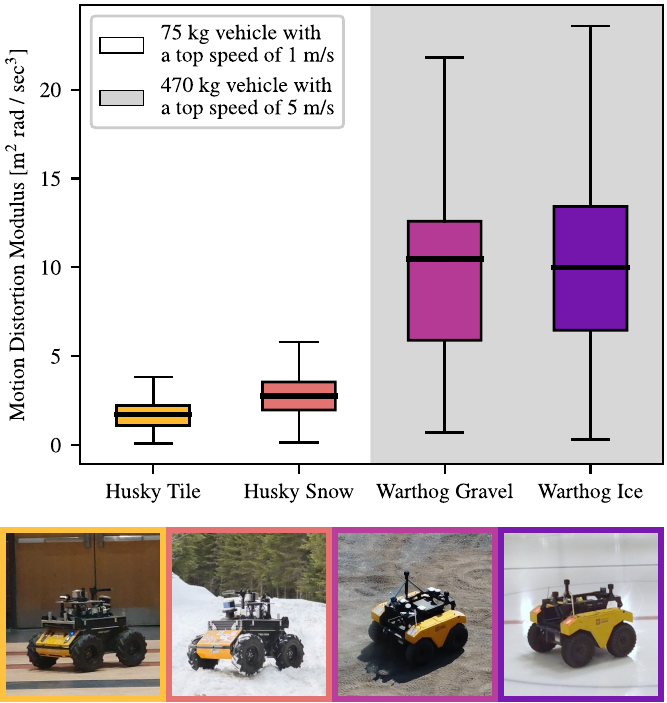}
    \caption{Motion distortion moduli were calculated on four datasets collected with the DRIVE protocol. 
    A \emph{Clearpath Husky A200} was used to record data separately on tile and on snow. 
    The two other datasets were collected with a \emph{Warthog} on gravel and on ice separately. 
    The \emph{Warthog} weighs \SI{470}{kg} and has a maximum speed of \SI{5}{m/s} while the \emph{Husky} has a maximum speed of \SI{1}{m/s} and weighs \SI{75}{kg}.}
    \label{fig:metrics}
\end{figure}

The resulting quantitative motion distortion metric for every time step of the dataset has been computed, and the results are presented in  \autoref{fig:metrics}. 
As supposed, the dataset made with the lightest vehicle going at the lowest speed on the simplest terrain (tile) has the smallest  motion distortion modulus median (\SI{1.716}{m^{2} rad/sec^{3}}). 
The motion distortion modulus median increases with the terrain complexity, as shown by the motion distortion modulus median of the Husky on snow (\SI{2.76}{m^{2} rad/sec^{3}}), which is 1.6 times bigger than the median obtained on tile.
The motion distortion modulus median also increases with the internal motion distortion features.  
The \emph{Warthog} is 6.2 times heavier and 5 times faster than the \emph{Husky}, and its dataset on ice has a motion distortion modulus median approximately 3.6 times greater than the motion distortion modulus median of the \emph{Husky} on snow. 

One limitation of the current metric is that it does not account for the error of transitory motion. 
Therefore, highly dynamic terrain like ice could have a similar motion modulus median to less dynamically challenging terrain like gravel. 
This phenomenon can be seen in \autoref{fig:metrics} where the motion distortion modulus median of the \emph{Warthog} on ice dataset is \SI{5}{\%} lower than the motion distortion modulus median of the \emph{Warthog} on gravel dataset.
Ice is more dynamically challenging than gravel because of a lower friction coefficient, lower traction force, and a longer transitory motion than gravel.

\section{Lessons learned}
\label{sec:lessons}
Over the last few years, we have focused on the problem of characterizing~\ac{UGV} motion in off-road and remote environments.
Notably, we quantified the behavior difference between~\ac{SSMR} navigation on concrete and snow-covered terrain through~\SI{2}{\kilo\meter} of driving data~\citep{Baril2020}.
To facilitate~\ac{UGV} behavior characterization, we recently released the~\ac{DRIVE} protocol and used it to gather~\SI{7}{\kilo\meter} and~\SI{1.8}{\hour} of driving data over three distinct UGVs and four terrain types~\citep{Baril2024}.\footnote{\url{https://github.com/norlab-ulaval/DRIVE}}
We also successfully completed~\SI{18.8}{\kilo\meter} of autonomous navigation in a subarctic forest during wintertime, a deployment through which we had to account for the high slip induced by snow-covered terrain~\citep{Baril2022}.
In this section, we provide a list of lessons learned,  with the intent of helping fellow field roboticists save time and bypass some mistakes we made in previous deployments.
All of these are listed and described below:
\begin{itemize}
    \item Controllers that require terrain characterization tend to perform better but need to be re-trained for any novel terrain conditions, which is costly in human time and energy.
    \item The energy and time budget allowed for motion characterization must be minimized for operation in remote areas, as both are crucial resources and any~\ac{UGV} breakdown has significant consequences. 
    Disaster response, forestry, and underground operations, as well as deployments in polar regions are examples of situations where these factors are critical. 
    \item The temperature and ground properties have a significant impact on the energy consumption of~\acp{UGV}. 
    For example, snow traction properties vary significantly with respect to temperature, sun illumination and humidity~\citep{Fierz2009}.
    For example, we observed an unexpected battery outage when executing the~\ac{DRIVE} protocol on wet snow.
    Thus, motion characterization protocols must be adapted for these features. 
    \item When recording data for model identification, acceleration ramps are critical to minimize premature mechanical wear and tear. 
    However, these ramps limited the range of internal features that were observed in our experiments.
    Removing them would allow the recording of extreme motions, at the cost of accelerated hardware wear and tear.
    \item Terrain properties are a continuous spectrum, multi-dimensional, and dependent on meteorological conditions. 
    Thus, friction on a terrain patch might differ over a single day and even more throughout the year. 
    This variation is especially true for winter navigation.
    \item The current lack of a database containing all the different motion distortion features makes it hard to validate motion models for all applications.
\end{itemize}

\section{Conclusion}
In this work, we investigate the problem of comparing various~\ac{UGV} deployments and the high variance in internal and external motion distortion features.
We propose a terrain-agnostic metric to estimate the difficulty of a motion dataset.
We show that this metric increases with the intensity of both internal and external motion distortion features.
However, it currently does not capture the impact of transitory motion caused by low terrain friction and aggressive driving.
Future work includes extending this metric to account for transitory motion distortion and using it to normalize the difficulty of motion datasets.
We also plan to generate a large collaborative dataset concatenating experiments with vehicles covering the entire spectrum of motion distortion features.

\printbibliography

\end{document}